# Graph Regularized Nonnegative Latent Factor Analysis Model for Temporal Link Prediction in Cryptocurrency Transaction Networks

Zhou Yue, Liu ZhiGang, and Yuan Ye

*Abstract*—With the development of blockchain technology, the cryptocurrency based on blockchain technology is becoming more and more popular. This gave birth to a huge cryptocurrency transaction network has received widespread attention. Link prediction learning structure of network is helpful to understand the mechanism of network, so it is also widely studied in cryptocurrency network. However, the dynamics of cryptocurrency transaction networks have been neglected in the past researches. We use graph regularized method to link past transaction records with future transactions. Based on this, we propose a single latent factor-dependent, non-negative, multiplicative and graph regularized-incorporated update (SLF-NMGRU) algorithm and further propose graph regularized nonnegative latent factor analysis (GrNLFA) model. Finally, experiments on a real cryptocurrency transaction network show that the proposed method improves both the accuracy and the computational efficiency.

## I. INTRODUCTION

With the wide application of blockchain technology, cryptocurrency has been widely considered as a new electronic alternative to exchange currency [7-9]. Different from the traditional currency needs the central authority to supervise the transaction, the cryptocurrency is maintained by the distributed consensus to ensure the efficiency of cryptocurrency transactions [1-3]. More importantly, the blockchain technology effectively records both parties' transactions in a verifiable and permanent manner, and the transaction records are not allowed to be modified. Therefore, cryptocurrencies provide great convenience for data mining and analysis in this field, they can publicly access the growing list of records stored in the chain, this list contains a wealth of information as well as a complete record of financial transactions [1-3]. Therefore, the formation of a huge cryptocurrency transaction network has been widely concerned [1-9].

Based on the transaction records of cryptocurrency, a large transaction network is constructed. Noted on that node in the network are abstracted from objects in the cryptocurrency system, such as accounts, smart contracts, and entities, while edges are abstracted from their relationships. The link prediction plays a vital role in the analysis of cryptocurrency transaction network, because it is helpful to the construction and analysis of the whole system [5, 6]. Muzammal *et al.* [10] decomposed the training graph into multiple subgraphs and then used probabilistic matrix decomposition (PMF) and Bayesian PMF model to extract hidden features in the subgraphs. Meo *et al.* [11] propose a the pairwise trust prediction through matrix factorization (PTP-MF) algorithm for pairwise trust prediction, the algorithm also combines the deviation of trustor and trustee behavior to predict the strength of trust and distrust relationship of users in cryptocurrency transaction network. However, they ignore that cryptocurrency transaction data are non-negative. In order to ensure non-negative data, Yu *et al.* [12] propose a double non-negative matrix factorization (DouNMF) model, which integrates degree information of nodes into nonnegative matrix factorization processes to obtain a more accurate node transfer matrix. Wang *et al.* [13] proposed a regularized convex nonnegative matrix factorization model (RC-NMF) that, by introducing graph regularization into convex nonnegative matrix factorization, could simultaneously constrain nodes with positive links to enter into the same community, it also constrains nodes with negative links to enter different communities.

However, the above method is inefficient because it needs to operate a huge complete matrix in computation and storage. In order to improve the computational efficiency, Luo *et al.* [14-16] proposed a single latent factor-dependent, non-negative, multiplicative and update (SLF-NMU) algorithm, which only depends on the known data in the network. Based on this algorithm, they further proposed a nonnegative latent factor analysis (NLFA) model. The computing cost of NLFA is linear only with the count of known items in its network, while the storage cost is linear only with the sum of its user count and item count.

However, the above method only applies to static cryptocurrency transaction network. In the real world, cryptocurrency transaction networks are dynamic, which means that the network structure evolves at different times. Analysis of the dynamic flow of cryptocurrency, which is critical to the analysis of the entire network. Therefore, compared with the static network, the dynamic

✧ Y Zhou and Z.G Liu are with the School of Computer Science and Technology, Chongqing University of Posts and Telecommunications, Chongqing 400065, and also with the Chongqing Key Laboratory of Big Data and Intelligent Computing, Chongqing Institute of Green and Intelligent Technology, Chinese Academy of Sciences, Chongqing 400714, China, and with the Chongqing School, University of Chinese Academy of Sciences, Chongqing 400714, China (e-mail: yueyzhou@outlook.com, liuzhigang @cigit.ac.cn, haowu@cqupt.edu.cn).

✧ Y. Yuan is with the College of Computer and Information Science, Southwest University, Chongqing 400715, China (e-mail: yuanyekl@gmail.com).

network is more accurate in the representation of complex systems. Therefore, temporal link prediction has been widely concerned.

Therefore, this paper focuses on the temporal link prediction in cryptocurrency transaction network. In this paper, we use graph regularized method to link past transaction records with future transactions. Based on this, we propose a single latent factor-dependent, non-negative, multiplicative and graph regularized-incorporated update (SLF-NMGRU) algorithm. With it, we further graph regularized nonnegative latent factor analysis (GrNLFA) model. The main contributions of this paper are as follows:

a) A GrNLFA model, which combines graph regularized and NLFA; and

b) Compared with other methods, this method greatly improves the computational efficiency without loss of precision. Experimental results on a real cryptocurrency transaction network show the superiority of the proposed algorithm.

Section II states the preliminaries. Section III presents the methods. Section IV conducts the empirical studies. Finally, Section V concludes this paper.

## II. PRELIMINARIES

### A. Problem Formulation

The graph $G$ is used to describe the cryptocurrency transaction network, which is defined as follows:

**Definition 1**: Let $G^t=(U^t, S^t\ R^t)^{\{t\in(1,...T)\}}$ denotes the cryptocurrency transaction network, where $U$ and $S$ represent the set of sender and receiver nodes in cryptocurrency transaction at time $t$, respectively. $R$ is the set of edges at time $t$, and $T$ is the time span. In particular, $G=\{G^1, G^2,…, G^T\}$, where $G^t$ represents the network at time $t$ with sender set $U$, receiver set $S$ and edge set $R$.

**Definition 2**: Let edge set $R^{|U|\times|S|}$ be a matrix in which each entry $r_{i,j}$ represents the relationship between $i\in U$ and $j\in S$; $\Lambda$ and $\Gamma$ be known and unknown entry sets of $R$.

It is important to note that $K$ is the dimension of latent factor (LF) spaces, $X$ and $Y$ are defined as the LF matrices reflecting characteristics of $U$ and $S$ represented by data in $\Lambda$.

### B. Non-negative Latent Factor Analysis Model

In order to extract non-negative LFs from known sets of data, a NLFA model uses the European distance to construct the loss function [14, 15, 18, 19]:

$$O^{NLF}(X,Y) = \|R - XY^T\|_F^2 = \sum_{r_{ij}\in\Lambda}\left(r_{ij} - \sum_{k=1}^{K} x_{ik} y_{jk}\right)^2 \quad (1)$$

$$s.t.\ \forall i\in U, j\in S, k\in\{1,2,...,K\}: x_{ik}\geq 0, y_{jk}\geq 0$$

The NLFA model uses the SLF-NMU algorithm to guarantee the nonnegativity of the algorithm. The updated rule is as follows:

$$O^{NLF}(X,Y) \stackrel{SLF-NMU}{\Rightarrow}$$

$$\begin{cases} x_{ik}^{d+1} \rightarrow x_{ik}^d \left(\sum_{j\in\Lambda(i)} y_{jk} r_{ij} \Big/ \sum_{j\in\Lambda(i)} y_{jk} \hat{r}_{ij}\right) \\ y_{jk}^{d+1} \rightarrow y_{jk}^d \left(\sum_{i\in\Lambda(j)} r_{ij} x_{ik} \Big/ \sum_{i\in\Lambda(j)} \hat{r}_{ij} x_{ik}\right) \end{cases} \quad (2)$$

where $d$ denotes the number of iterations. With (2), the original NLFA model is complete, and it can be applied to a variety of data analysis tasks [28, 29, 32].

## III. METHODS

### A. Objective Function

The graph regularization method has an effective strategy [21-24]. Its core idea is given data matrix $R=[R_1,…,R_n]$, if the two objects $R_i$ and $R_j$ are close in the intrinsic geometry of the data distribution, then the feature vectors $Y_i$ and $Y_j$ (the $i$th and $j$th column of the feature matrix $Y$) are also close for the new basis. Therefore, by combining the (1) with the graph regularized method, we obtain the regularized term for $Y$ as follows:

$$O^{Gr}(Y) = \alpha \|y_{j,\cdot} - y_{l,\cdot}\|^2 w_{jl} \quad (3)$$

where $w_{jl}$ is the weight matrix used to measure the compactness between two points $y_{j,\cdot}$ and $y_{l,\cdot}$, $\alpha$ is the regularization parameter controls the smoothness of the new representation. However, there are multiple networks in cryptocurrency transaction networks.

Based on previous studies [20, 21], it is assume that the graph regularization from different historical transaction networks is located in the convex hull of a previously given candidate manifold. This assumption implies that the search space for possible graph Laplace operator is a linear combination of $T$-1 regularization terms. Therefore, we control the relative importance of $G^t$ by the parameter $\theta$, and we obtain the following regularization terms:

$$O^{Gr}(Y,\theta) = \alpha \sum_{t=1}^{T-1} \theta^{T-t} \|y_{j,\cdot} - y_{l,\cdot}\|^2 w_{jl}^t \quad (4)$$

By combining (1) and (4), we complete the final objective function as follows:

$$O^{GrNLF}(X,Y) = O^{NLF}(X,Y) + O^{Gr}(Y,\theta)$$

$$= \sum_{r_{ij} \in \Lambda}\left(\left(r_{ij} - \sum_{k=1}^{K} x_{ik} y_{jk}\right)^2 + \alpha \sum_{t=1}^{T-1}\sum_{l=1}^{N}\sum_{k=1}^{K}\left(\theta^{T-t}\left(y_{jk} - y_{lk}\right)^2 w_{jl}^t\right)\right) \quad (5)$$

$$s.t. \quad \forall i \in M, j,l \in N, k \in \{1,2,...,K\}, t \in \{1,2,..,T\}: x_{ik} \geq 0, y_{jk} \geq 0$$

Based on (5), we achieve the objective function for a GrNLFA.

*B. SLF-NMGRU Algorithm*

To guarantee the nonnegativity of matrices $X$ and $Y$ in (5), we use Lagrange method. To constrain $x_{ik} \geq 0$ and $y_{jk} \geq 0$, we construct Lagrange multiplier $\psi_{ik}$ and $\phi_{jk}$, the Lagrange $\mathcal{L}$ of (5) as follows:

$$\mathcal{L}^{GrNLF} = O^{GrNLF} + Tr(\Psi X^T) + Tr(\Phi Y^T)$$

$$= O^{GrNLF} + \sum_{j \in \Lambda(i)}\sum_{k=1}^{K}\psi_{ik}x_{ik} + \sum_{i \in \Lambda(j)}\sum_{k=1}^{K}\phi_{jk}y_{jk} \quad (6)$$

The partial derivative of $\mathcal{L}$ with respect to $X$ and $Y$ is as follows:

$$\begin{cases} \dfrac{\partial \mathcal{L}^{GrNLF}}{\partial X} = \sum_{j \in \Lambda(i)}\left(-y_{jk}(r_{ij} - \hat{r}_{ij})\right) + \psi_{ik} \\ \dfrac{\partial \mathcal{L}^{GrNLF}}{\partial Y} = \sum_{i \in \Lambda(j)}\left(\begin{array}{l}\left(-x_{ik}(r_{ij} - \hat{r}_{ij})\right) + \\ \alpha\sum_{t=1}^{T-1}\sum_{l=1}^{N}\theta^{T-t}\left(w_{jl}^t(y_{jk} - y_{lk})\right)\end{array}\right) + \phi_{jk} \end{cases} \quad (7)$$

According to the Karush-Kuhn-Tucker conditions $\psi_{ik}x_{ik}=0$ and $\phi_{jk}y_{jk}=0$, the equations for $X$ and $Y$ are obtained as follows:

$$\begin{cases} x_{ik}\sum_{j \in \Lambda(i)}\left(y_{jk}\hat{r}_{ij} - y_{jk}r_{ij}\right) + \psi_{ik}x_{ik} = 0 \\ y_{jk}\sum_{i \in \Lambda(j)}\left(\begin{array}{l}\left(\hat{r}_{ij}x_{ik} - r_{ij}x_{ik}\right) + \\ \alpha\sum_{t=1}^{T-1}\sum_{l=1}^{N}\theta^{T-t}\left(w_{jl}^t(y_{jk} - y_{lk})\right)\end{array}\right) + \phi_{jk}y_{jk} = 0 \end{cases} \quad (8)$$

According to (8), we get the final update rule as follows:

$$O^{GrNLF}(X,Y) \stackrel{SLF-NMGRU}{\Rightarrow} \begin{cases} x_{ik}^{d+1} \rightarrow x_{ik}^d \dfrac{\sum_{j \in \Lambda(i)} y_{jk} r_{ij}}{\sum_{j \in \Lambda(i)} y_{jk}\hat{r}_{ij}} \\ y_{jk}^{d+1} \rightarrow y_{jk}^d \dfrac{\sum_{i \in \Lambda(j)} r_{ij}x_{ik} + \alpha|\Lambda(j)|\sum_{t=1}^{T-1}\sum_{l=1}^{N}\theta^{T-t}w_{jl}^t y_{lk}}{\sum_{i \in \Lambda(j)} \hat{r}_{ij}x_{ik} + \alpha|\Lambda(j)|\sum_{t=1}^{T-1}\sum_{l=1}^{N}\theta^{T-t}w_{jl}^t y_{jk}} \end{cases} \quad (9)$$

According (9), we achieve the SLF-NMGRU algorithm for GrNLFA model.

*C. Algorithm Design and Alalysis*

On the basis of Section III (A) and (B), we design the algorithm of GrNLFA model, as shown in the algorithm GrNLFA. Note that the algorithm GrNLFA summarizes the time cost of each step. According to these summaries, the total time cost is:

$$T_{GrNLFA} \approx \Theta(n \times |\Lambda| \times d) \quad (10)$$

where the condition of (10) is that $|\Lambda| \gg \max\{|U|,|S|\}$, used in a variety of applications to drop lower-order-complexity terms and constants.

The storage cost of GrNLFA relies on $\Lambda$, $U$, $S$, $d$ related auxiliary arrays, yielding:

$$S_{GrNlFA} \approx \Theta|\Lambda| + (|U| + |S|) \times d. \quad (11)$$

which is linear with the involved entity count.

| **Algorithm GrNLF** |
|---|
| **Input:** *G*: cryptocurrency transaction networks |
| *K*: dimension of LF spaces |
| *α*: regularization parameter |

| $\theta$: time slice parameter | |
|---|---|
| **Operation** | **Cost** |
| **initialize** $X^{|U|\times d}$, $Y^{|S|\times d}$ | $\Theta((|U|+|S|)\times d)$ |
| **initialize** $\alpha$, $\theta$, $d$=0, *Max-training-round=N* | $\Theta(1)$ |
| **while not** converge **and** $n\leq N$ **do** | $\times n$ |
|   **for each** $r_{ij}$ **in** $\Lambda$ | $\times|\Lambda|$ |
|     Update $x_{i,\cdot}$ and $y_{j,\cdot}$ with (9) | $\Theta(d)$ |
|   **end for** | - |
|   $n=n+1$ | $\Theta(1)$ |
| **end while** | - |
| **Output:** *X, Y* | |

IV. EXPERIMENTAL RESULTS AND ANALYSIS

*A. General Settings*

**Evaluation Metrics.** In this study, the root mean squared error (RMSE) and mean absolute error (MAE) are picked up as an evaluation metric [18, 19, 30-35, 42-49]:

$$RMSE = \sqrt{\left(\sum_{r_{u,v}\in K} \left|r_{u,v} - \hat{r}_{u,v}\right|^2\right)\bigg/|K|},$$

$$MAE = \left(\sum_{r_{u,v}\in K} \left|r_{u,v} - \hat{r}_{u,v}\right|_{abs}\right)\bigg/|K|; \quad (12)$$

where K is the validation set disjoint with the training set $\Lambda$ and the testing set $\Gamma$, and $\hat{r}_{u,v}$ denotes the estimation generated by a tested model for $\forall r_{u,v}\in K$, respectively.

**Datasets.** As shown in Table I, two real cryptocurrency transaction networks were used in our experiment. For temporal link prediction, we divide 1 to *T*-2 as training set, *T*-1 as verification set, and the last time slice as test set.

TABLE I. Experimental dataset details.

| No. | Name | Sender | Receive | Time | Edge |
|---|---|---|---|---|---|
| **D1** | Ethereum transaction network [28] | 8,626 | 2,904 | 1,014 | 203,921 |
| **D2** | Bitcoin transaction network [11] | 6,351 | 7,784 | 237 | 50,178 |

**Model Settings.** Table II summarizes details of compared models. In addition, the learning process of a tested model terminates if i) it converges, i.e., its error difference between two consecutive epochs becomes smaller than $10^{-5}$, or ii) its iteration count reaches a given threshold, i.e., 1000.

TABLE II. Compared Models.

| No. | Name | Description |
|---|---|---|
| **M1** | NMF | The classical NMF model approximates the original matrix by decomposing the product of two nonnegative matrices.[36] |
| **M2** | GrNMF | A GrNMF model uses graph regularization to regularize the historical network. [21] |
| **M3** | GrNLF | A GNALF model relying on SLF-NMGRU proposed in Section III (B) |

*B. Parameter Sensitivity*

As discussed in Section III, this section discusses the effect of the time slice parameter $\theta$ on the GRNLF. Fig. 1 depicts the effect of $\theta$ on GRNLF data at D1 and D2. From these results, we have the following findings:

**a) For SLF-NMGRU as a learning algorithm, both RMSE and MAE of M3 are sensitive to $\theta$.** For instance, as shown in Fig. 1(a), RMSE of M4 with $\theta=2^{-1}$, $2^{-2}$, $2^{-3}$, $2^{-4}$, and $2^{-5}$ is 0.6452, 0.6408, 0.6355, 0.6457, 0.6481, respectively. RMAE of $\theta=2^{-3}$ increases 3.64%, 0.83%, 1.57% and 1.94%, respectively. Similar findings are found in the remaining data, as shown in Fig. 1(b)-(d).

**b) Small $\theta$ affects the prediction accuracy of M3.** For example, as depicted in Fig.1 (c), compared with $\theta=2^{-5}$ and $\theta=2^{-2}$, the prediction precision lost 6.03%. Similar findings are found in the remaining data, as shown in Fig. 1(a), (b) and (d)In this part of experiments, we call this manually-tuned ADNLF model as the ADNLF-M. Figs. 1 depicts its RMSE as $\alpha$ and $\beta$ vary. Table II records ADNLF-M's lowest RMSE with optimal $\alpha$ and $\beta$ versus its RMSE as $\alpha=\beta=1$. From them, we have the following findings:

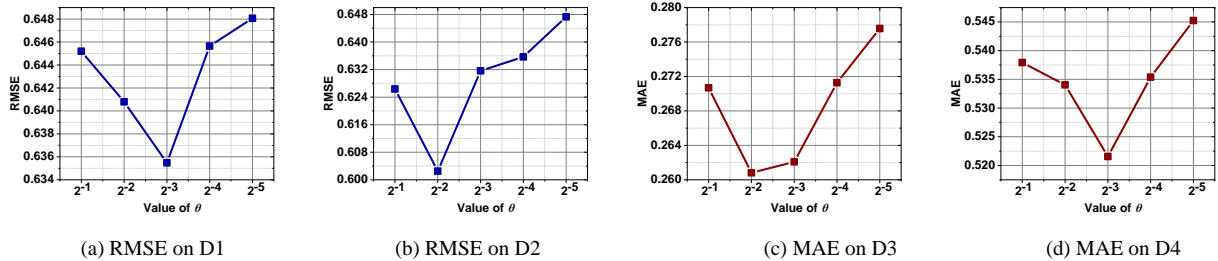

(a) RMSE on D1    (b) RMSE on D2    (c) MAE on D3    (d) MAE on D4

Fig.1. Effects of $\theta$ in M3 on D1 and D2.

*C. Comparison Results*

In this part, we aim to clarify their performance by comparing the proposed M3 model with other models. Figs. 2 and 3 depicts the training process from M1 to M3. Table III summarizes the lowest RMAE and MAE, the rounds of iterations, and the total training time. From these results, we have a discovery:

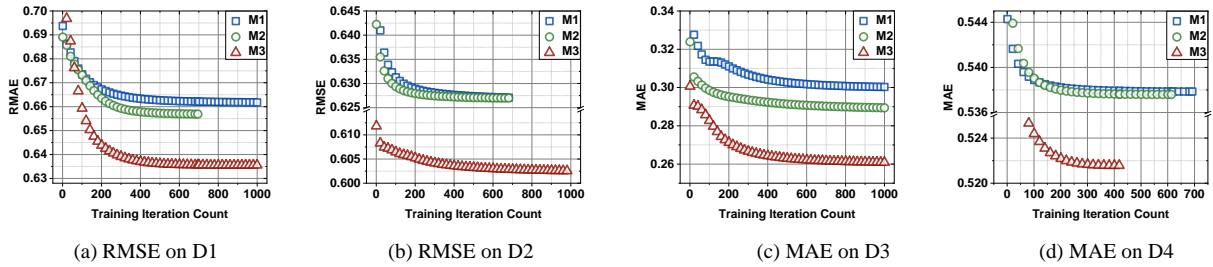

(a) RMSE on D1    (b) RMSE on D2    (c) MAE on D3    (d) MAE on D4

Fig.2. Comparison of M1-M3 on D1 and D2.

a) The GrNLFA model has higher prediction accuracy. As shown in Fig. 2(a), the RMSE of M1-M3 is 0.3003, 0.2894, 0.2608, respectively. The M3' RMSE is 3.97% and 3.26% higher than of M1's RMSE and M2'RMSE, respectively. The results are similar for other datasets, as shown in figs. 2 (b) and 3.

b) The GrNLFA model greatly improves computational efficiency. Due to M3 only depends on the known value of the historical transaction data, it avoids the full matrix operation and greatly improves the storage efficiency and the computational efficiency of the model. Thus, as shown in Table III, the time consumption for M1 to M3 is increased by 99.9% and 99.9% for 61, 078s, 1,093,892s, and 18s, and M2, respectively. The results are similar for other data sets, as shown in Table III.

TABLE V. Lowest RMSE and their corresponding total time cost (Secs).

|  | Case | RMSE | MAE | Round (RMSE) | Round (MAE) | Total Time/s (RMSE) | Total Time/s (MAE) |
|---|---|---|---|---|---|---|---|
| **D1** | M1: | 0.6619 | 0.3003 | 1,000 | 1,000 | 61,078 | 61,078 |
|  | M2: | 0.6569 | 0.2894 | **695** | 1,000 | 1,093,892 | 1,573,946 |
|  | M3: | **0.6354** | **0.2608** | 1,000 | 1,000 | **18** | **18** |
| **D4** | M1 | 0.6271 | 0.5379 | 695 | 691 | 19,923 | 19,894 |
|  | M2 | 0.6269 | 0.5375 | 692 | 614 | 136,389 | 121,015 |
|  | M3: | **0.6025** | **0.5216** | **265** | 419 | **18** | **7** |

## V. CONCLUSIONS

In this paper, SLF-NMGRU algorithm and GrNLFA model are proposed. Experimental results in real cryptocurrency transaction networks show the superiority of the proposed model. In addition, considering future research, the following directions deserve further discussion:

a) The tensor method can be used to express the temporal characteristics of networks [12, 17, 20, 27, 37, 38, 50].

b) The performance of the model can be further optimized, and effective optimization methods such as second-order method [26, 40, 41] and momentum method [28, 39, 47] can be studied.

REFERENCES

[1] Y. Liu, F. R. Yu, X. Li, H. Ji and V. C. M. Leung, "Blockchain and Machine Learning for Communications and Networking Systems," IEEE Communications Surveys Tutorials, vol. 22, no. 2, pp. 1392-1431, 2020.
[2] Y. Lu, "The blockchain: State-of-the-art and research challenges," Journal of Industrial Information Integration, vol. 15, pp. 80–90, Sep. 2019.
[3] M. H. u. Rehman, K. Salah, E. Damiani and D. Svetinovic, "Trust in Blockchain Cryptocurrency Ecosystem," IEEE Trans. on Engineering Management, vol. 67, no. 4, pp. 1196-1212, Nov. 2020.
[4] E. Livieris, N. Kiriakidou, S. Stavroyiannis, and P. Pintelas, "An advanced CNN-LSTM model for cryptocurrency forecasting," Electronics, vol. 10, no. 3, p. 287, Jan. 2021
[5] W. Chen, J. Wu, Z. Zheng, C. Chen, and Y. Zhou, ''Market manipulation of bitcoin: Evidence from mining the Mt. Gox transaction network,'' in Proc. IEEE Conf. Computer Communications (INFOCOM), pp. 964–972, Apr. 2019.

[6] Q. Yuan, B. Huang, J. Zhang, J. Wu, H. Zhang and X. Zhang, "Detecting Phishing Scams on Ethereum Based on Transaction Records," 2020 IEEE Int. Symposium on Circuits and Systems (ISCAS), pp. 1-5, Sept. 2020.
[7] M. C. K. Khalilov and A. Levi, "A survey on anonymity and privacy in Bitcoin-like digital cash systems," IEEE Communications Surveys Tutorials, vol. 20, no. 3, pp. 2543–2585, Mar. 2018.
[8] F. Tschorsch and B. Scheuermann, "Bitcoin and beyond: A technical survey on decentralized digital currencies," IEEE Communications Surveys Tutorials, vol. 18, no. 3, pp. 2084–2123, Mar. 2016.
[9] M. Conti, E. S. Kumar, C. Lal, and S. Ruj, "A survey on security and privacy issues of bitcoin," IEEE Communications Surveys Tutorials, vol. 20, no. 4, pp. 3416–3452, Apr. 2018.
[10] M. Muzammal, F. Abbasi, Q. Qu, R. Talat and J. P. Fan, "A decentralised approach for link inference in large signed graphs," Future Generation Computer Systems, vol. 102, pp. 827-837, Jan. 2020.
[11] P. Meo, "Trust Prediction via Matrix Factorisation," ACM Trans. on Internet Technology, vol. 19, no. 44, pp. 1-20, Nov. 2019.
[12] W. Yu, R. Mu, Y. Sun, X. Chen, W. Wang and H. Wu, "A Double Non-Negative Matrix Factorization Model for Signed Network Analysis," in 2019 IEEE Int. Conf. on Parallel & Distributed Processing with Applications, Big Data & Cloud Computing, Sustainable Computing & Communications, Social Computing & Networking (ISPA/BDCloud/SocialCom/SustainCom), pp. 936-943, Dec. 2019.
[13] J. Wang and R. J. Mu, "A Regularized Convex Nonnegative Matrix Factorization Model for signed network analysis," Social Network Analysis and Mining, vol. 11, no. 7, Jan. 2021.
[14] X. Luo, M.-C. Zhou, Y.-N. Xia, Q.-S. Zhu, A. C. Ammari, and A. Alabdulwahab, "Generating highly accurate predictions for missing QoS data via aggregating nonnegative latent factor models," IEEE Trans. On Neural Networks and Learning Systems, vol. 27, no. 3, pp. 524-537, 2016.
[15] X. Luo, M.-C. Zhou, Y.-N. Xia, and Q.-S. Zhu, "An efficient non-negative matrix-factorization-based approach to collaborative filtering for recommender systems," IEEE Trans. on Industrial Informatics, vol. 10, no. 2, pp. 1273-1284, 2014.
[16] Z. Liu, X. Luo, and Z. Wang, "Convergence Analysis of Single Latent Factor-dependent, Non-negative and Multiplicative Update-based Non-negative Latent Factor Models," IEEE Trans. on Neural Networks and Learning Systems, DOI: 10.1109/TNNLS.2020.2990990.
[17] H. Wu, X. Luo and M. Zhou, "Advancing non-negative latent factorization of tensors with diversified regularizations," IEEE Trans. on Services Computing, DOI: 10.1109/TSC.2020.2988760.
[18] X. Zhou, M. Zhou, S. Li, and M. Shang, "An inherently non-negative latent factor model for high-dimensional and sparse matrices from industrial applications," IEEE Trans. Ind. Informat., vol. 14, no. 5, pp. 2011–2022, May 2018.
[19] X. Luo, M.-C. Zhou, S. Li, Z.-H. You, Y.-N. Xia, and Q.-S. Zhu, "A nonnegative latent factor model for large-scale sparse matrices in recommender systems via alternating direction method," IEEE Trans. Neural Netw. Learn. Syst., vol. 27, no. 3, pp. 579–592, Mar. 2016.
[20] D. M. Dunlavy, T. G. Kolda, and E. Acar, "Temporal link prediction using matrix and tensor factorizations," Acm Transactions on Knowledge Discovery from Data, vol. 5, no. 2, pp. 1–27, 2011.
[21] X. Ma, P. Sun, and Y. Wang, "Graph regularized nonnegative matrix factorization for temporal link prediction in dynamic networks," Physica A Statistical Mechanics & Its Applications, vol. 496, 2018.
[22] D. Cai, X. He, J. Han and T. S. Huang, "Graph Regularized Nonnegative Matrix Factorization for Data Representation," IEEE Trans. on Pattern Analysis and Machine Intelligence, vol. 33, no. 8, pp. 1548-1560, Aug. 2011.
[23] X. Li, G. Cui and Y. Dong, "Graph Regularized Non-Negative Low-Rank Matrix Factorization for Image Clustering," in IEEE Transactions on Cybernetics, vol. 47, no. 11, pp. 3840-3853, Nov. 2017.
[24] C. Leng, H. Zhang, G. Cai, I. Cheng and A. Basu, "Graph regularized Lp smooth non-negative matrix factorization for data representation," IEEE/CAA Journal of Automatica Sinica, vol. 6, no. 2, pp. 584-595, Mar. 2019.
[25] D. Lin, J. Wu, Q. Yuan, and Z. Zheng, "T-edge: Temporal weighted multidigraph embedding for Ethereum transaction network analysis," Frontiers in Physics, vol. 8, pp. 204, May. 2020.
[26] X. Luo, M. C. Zhou, S. Li, Y. N. Xia, Z. H. You, Q. S. Zhu, and H. Leung, "Incorporation of Efficient Second-order Solvers into Latent Factor Models for Accurate Prediction of Missing QoS Data," IEEE Trans. on Cybernetics, vol. 48, no. 4, pp. 1216-1228, April 2018.
[27] X. Luo, H. Wu, M. Zhou and H. Yuan, "Temporal pattern-aware QoS prediction via biased non-negative latent factorization of tensors," IEEE Trans. on Cybernetics, vol. 50, no. 5, pp. 1798-1809, May 2020.
[28] X. Luo, Z. Liu, S. Li, M. Shang and Z. Wang, "A Fast Non-Negative Latent Factor Model Based on Generalized Momentum Method," IEEE Trans. on Systems, Man, and Cybernetics: Systems, vol. 51, no. 1, pp. 610-620, Jan. 2021.
[29] X. Luo, M. Shang and S. Li, "Efficient Extraction of Non-negative Latent Factors from High-Dimensional and Sparse Matrices in Industrial Applications," in Proc. of 2016 IEEE 16th Int.Confer. on Data Mining (ICDM), 2016, pp. 311-319.
[30] D. Wu, Q. He, X. Luo, M. Shang, Y. He and G. Wang, "A Posterior-Neighborhood-Regularized Latent Factor Model for Highly Accurate Web Service QoS Prediction," IEEE Trans. on Services Computing, vol. 15, no. 2, pp. 793-805, Mar. 2022.
[31] L. Hu, X. Yuan, X. Liu, S. Xiong and X. Luo, "Efficiently Detecting Protein Complexes from Protein Interaction Networks via Alternating Direction Method of Multipliers," IEEE/ACM Trans. on Computational Biology and Bioinformatics, vol. 16, no. 6, pp. 1922-1935, 1 Nov. 2019.
[32] L. Xin, Y. Yuan, M. Zhou, Z. Liu and M. Shang, "Non-Negative Latent Factor Model Based on β-Divergence for Recommender Systems," IEEE Trans. on Systems, Man, and Cybernetics: Systems, vol. 51, no. 8, pp. 4612-4623, Aug. 2021.
[33] D. Wu, X. Luo, M. Shang, Y. He, G. Wang, and X. Wu, "A data-characteristic-aware latent factor model for web services QoS prediction," IEEE Trans. on Knowledge and Data Engineering, DOI:10.1109/TKDE.2020.3014302.
[34] X. Luo, Y. Zhou, Z. G. Liu, and M. C. Zhou, "Fast and Accurate Non-negative Latent Factor Analysis on High-dimensional and Sparse Matrices in Recommender Systems," IEEE Transactions on Knowledge and Data Engineering, DOI: 10.1109/TKDE.2021.3125252.
[35] X. Luo, Y. Zhou, Z. Liu, L. Hu and M. Zhou, "Generalized Nesterov's Acceleration-incorporated Non-negative and Adaptive Latent Factor Analysis," IEEE Trans. on Services Computing, doi: 10.1109/TSC.2021.3069108.
[36] D. Lee, "Learning the parts of objects with nonnegative matrix factorization," Nature, vol. 401, no. 6755, p. 788, 1999.
[37] X. Luo, H. Wu, Z. Wang, J. J. Wang, and D. Y. Meng, "A Novel Approach to Large-Scale Dynamically Weighted Directed Network Representation," IEEE Transactions on Pattern Analysis and Machine Intelligence, DOI: 10.1109/TPAMI.2021.313250.
[38] H. Wu, X. Luo, M. C. Zhou, M. J. Rawa, K. Sedraoui, and A. Albeshri, "A PID-incorporated Latent Factorization of Tensors Approach to Dynamically Weighted Directed Network Analysis," IEEE/CAA Journal of Automatica Sinica, DOI: 10.1109/JAS.2021.1004308.
[39] Y. R. Zhong, L. Jin, M. S. Shang, and X. Luo, "Momentum-incorporated Symmetric Non-negative Latent Factor Models," IEEE Transactions on Big Data, DOI 10.1109/TBDATA.2020.3012656.
[40] X. Luo, M.-C. Zhou, S. Li, Z.-H. You, Y.-N. Xia, Q.-S. Zhu, and H. Leung, "An efficient second-order approach to factorizing sparse matrices in recommender systems," IEEE Trans. on Industrial Informatics, vol. 11, no. 4, pp. 946 - 956, 2015.
[41] W. L. Li, Q. He, X. Luo, and Z. D. Wang, "Assimilating Second-Order Information for Building Non-Negative Latent Factor Analysis-Based Recommenders," IEEE Transactions on System Man Cybernetics: Systems, vol. 52, no. 1, pp. 485-497, 2021.
[42] D. Wu, Y. He, X. Luo, and M. C. Zhou, "A Latent Factor Analysis-based Approach to Online Sparse Streaming Feature Selection," *IEEE Transactions on System Man Cybernetics: Systems*, DOI: 10.1109/TSMC.2021.3096065.
[43] D. Wu, M. S. Shang, X. Luo, and Z. D. Wang, "An L1-and-L2-norm-oriented Latent Factor Model for Recommender Systems," *IEEE Transactions on Neural Networks and Learning Systems*, DOI: 10.1109/TNNLS.2021.3071392.

[44] M. S. Shang, Ye Yuan, X. Luo, and M. C. Zhou, "An α-β-divergence-generalized Recommender for Highly-accurate Predictions of Missing User Preferences," IEEE Transactions on Cybernetics, DOI: 10.1109/TCYB.2020.3026425.

[45] D. Wu and X. Luo, "Robust Latent Factor Analysis for Precise Representation of High-Dimensional and Sparse Data," IEEE/CAA Journal of Automatica Sinica, vol. 8, no. 4, pp. 796-805, Apr. 2021.

[46] Y. Yuan, Q. He, X. Luo, and M. S. Shang, "A Multilayered-and-Randomized Latent Factor Model for High-Dimensional and Sparse Matrices," IEEE Transactions on Big Data, DOI: 10.1109/TBDATA.2020.2988778.

[47] L. Hu, X. Pan, Z. Tan and X. Luo, "A Fast Fuzzy Clustering Algorithm for Complex Networks via a Generalized Momentum Method," *IEEE Trans. on Fuzzy Systems*, doi: 10.1109/TFUZZ.2021.3117442.

[48] X. Y. Shi, Q. He, X. Luo, Y. N. Bai, and M. S. Shang, "Large-scale and Scalable Latent Factor Analysis via Distributed Alternative Stochastic Gradient Descent for Recommender Systems," IEEE Transactions on Big Data, DOI: 10.1109/TBDATA.2020.2973141.

[49] X. Luo, Y. Yuan, S. L. Chen, N. Y. Zeng, and Z. D. Wang, "position-transitional particle swarm optimization-incorporated latent factor analysis," IEEE Trans. on Knowledge and Data Engineering, DOI:10.1109/TKDE.2020.3033324.

[50] X. Luo, H. Wu and Z. Li, "NeuLFT: A Novel Approach to Nonlinear Canonical Polyadic Decomposition on High-Dimensional Incomplete Tensors," *IEEE Trans. on Knowledge and Data Engineering*, doi: 10.1109/TKDE.2022.3176466.